\documentclass[conference]{IEEEtran}

\usepackage{cite}
\usepackage{amsmath,amssymb,amsfonts}
\usepackage{algorithm}
\usepackage{algorithmic}
\usepackage{graphicx}
\usepackage{textcomp}
\usepackage{xcolor}
\usepackage{comment}
\usepackage{hyperref}

\def\BibTeX{{\rm B\kern-.05em{\sc i\kern-.025em b}\kern-.08em
    T\kern-.1667em\lower.7ex\hbox{E}\kern-.125emX}}

\usepackage{caption}
\usepackage{subcaption}

\usepackage{anyfontsize}
\usepackage[T1]{fontenc}

\IEEEoverridecommandlockouts

\begin{document}

\title{Bayesian Structural Learning for an Improved Diagnosis of Cyber-Physical Systems}

    \author{\IEEEauthorblockN{Nicolas Olivain\textsuperscript{\textsection}}
    \IEEEauthorblockA{\textit{Mines Paristech - Paris Sciences \& Lettres} \\
    Paris, France \\
    nicolas.olivain@mines-paristech.fr}
		
    \and
    \IEEEauthorblockN{Philipp Tiefenbacher\textsuperscript{\textsection}, Jens Kohl\textsuperscript{\textsection}}
    \IEEEauthorblockA{\textit{Development Powertrain} \\
    \textit{BMW Group}\\
    Munich, Germany \\
    \{philipp.tiefenbacher, jens.kohl\}@bmw.com}
		}

\maketitle
\thispagestyle{plain}
\pagestyle{plain}

\begingroup\renewcommand\thefootnote{\textsection}
\footnoterule
\footnotetext{All authors contributed equally.}
\endgroup

\begin{abstract}
    The diagnosis of cyber-physical systems aims to detect faulty behaviour, its root cause and a mitigation or even prevention policy. 
    Therefore, diagnosis relies on a representation of the system's functional and faulty behaviour combined with observations of the system taken at runtime. 
    The main challenges are the time-intensive building of a model, possible state-explosion while searching for the root cause and interpretability of the results. In this paper we propose a scalable algorithm tackling these challenges. 
    We use a Bayesian network to learn a structured model automatically and optimise the model by a Genetic algorithm. 
    Our approach differs from existing work in two aspects: instead of selecting features prior to the analysis we learn a global representation using all available information which is then transformed to a smaller, label-specific one and we focus on interpretability to facilitate repairs. 
    The evaluation shows that our approach is able to learn a model with equal performance to state-of-the-art algorithms while giving better interpretability and having a reduced size.    
\end{abstract}


\section{Introduction} \label{sec:Intro}
\subsection{Diagnosis of cyber-physical systems} \label{subsec:challenges}

In cyber-physical systems (CPS, defined in e.g. \cite{rajkumar2010cyber}) mechanical, electrical and electronic components are combined with and controlled by software components to execute defined tasks. They are widely used in consumer electronics, Internet-of-Things devices, industrial plants, airplanes or vehicles.
Since the tasks assigned to a CPS can be safety-relevant and harmful to its environment, possible faulty behaviour has to be detected, the fault's root cause has to be inferred and failures have to be prevented or mitigated. Furthermore, details about the fault's root cause are necessary for a system remedy via repair. These are the central tasks of diagnosis.

For this paper we use the terminology as introduced by \cite{avizienis2004basic} wherein a \textit{failure} is defined as an event that occurs when the delivered service deviates from correct service. An \textit{error} is the part of the total state of the system that may lead to its subsequent service failure. The adjudged or hypothesized cause of an error is called a \textit{fault}. These terms have a causal chain as in that a fault causes an error which then can lead to a failure.

The diagnosis of CPS consists of several challenges regarding \textit{building or learning the diagnosis' model} given the CPS' \textit{interconnections and dependencies} and \textit{imbalanced data} from different \textit{data sources}, running the diagnosis model under \textit{hard real-time constraints} delivering \textit{fast inference of the root cause(s)} for failure mitigation or prevention and a \textit{focused repair}.

\textit{Building the diagnosis model} for a CPS is a complex and especially time-consuming process due to lots of functional dependencies and interconnections between the components of a CPS as well as with other interacting CPS. Additionally, CPS components can have fault dependencies while having no functional dependencies (e.g., blocking a shared communication device thus causing a time out). Thus, a lot of these dependencies are `learned' by trial-and-error.

Since manually building a diagnosis model requires lots of effort, using supervised learning algorithms for \textit{learning a diagnosis model} seems promising. However, using such approaches is impeded since the available data is highly imbalanced. Given that a vast majority of CPS operate according to their specification, supervised learning approaches tend to regard the few faulty CPS as noise or outliers.
Additionally, the available observations of a faulty CPS are imbalanced as well. While CPS typically sample hundreds or more observations of their environment and internal behaviour with rates of 1000 Hertz or more, their purely mechanical parts cannot be sampled due limited insight.

Hence, to increase the amount of available data for mechanical parts other observations along their whole life-cycle have to be collected, such as the production data of their individual parts, their assembly into a CPS and -if available- customer feedback. Since these data points belong to different \textit{data sources}, they need to be encoded as observations.

For safety-relevant domains, the diagnosis of CPS has to adhere to \textit{(hard) real-time constraints} meaning faulty behaviour has to be detected (almost) instantly followed by searching for diagnosis candidates fitting the model and the observations. For complex CPS this can result in an exponential state space impeding a \textit{fast inference of the root cause}.

Since most CPS provide only limited direct access or insight, a \textit{focused repair} to remedy the CPS depends on concise diagnosis results. Concise, however, depends: while domain experts need extensive information about the root cause - especially during development of a system - service people or end-users prefer understandable results to support a focused repair.

Finally, the CPS' diagnosis model needs to be maintained over the whole product lifecycle which typically spans several years.

\subsection{Overview of our contribution}

The contribution of this paper is a scalable methodology able to cope with the detailed challenges. We use a Bayesian Network to automatically learn a feature ensemble explaining the functional and fault relationships as well as dependencies of the CPS. This learned representation is optimized by using a genetic algorithm, thus returning a reduced diagnosis model with equal accuracy as other comparable algorithms while having a better interpretability. A huge benefit of our approach is that it works with imbalanced data -as quite common for CPS- by using information-based criteria as an evaluation metric both when building the network and in the genetic algorithm's fitness function. The increased interpretability is especially helpful for domain experts in the development phases and for non-domain experts during repairs. 

Furthermore, the Markov-separated model is easily maintainable by domain experts. They can integrate further knowledge from different data sources as Boolean formulas into the model and transfer separated parts of the model to similar CPS or domains without many changes.

Finally, having a reduced model helps avoiding the state explosion problem for the inference. This allows us to deploy the model on CPS with limited computing resources.

\subsection{Outline}
In this section we described our problem domain, detailed its main challenges and outline our contribution. In the next section, Sec. \ref{sec:Related}, we discuss related work. Sec. \ref{sec:contribution} details our contribution and its benefits to our problem domain. In Sec. \ref{sec:evaluation} we show an evaluation of our methodology on two different data sets, a medical dataset and one for an automotive component. Finally, Sec. \ref{sec:conclusion} concludes this paper and shows possible future work.

\section{Discussion of related Work} \label{sec:Related}

\textbf{Modeling Cyber-physical systems}. Several model-based methodologies have been used for a representation of a CPS' functional behaviour~\cite{buck1994ptolemy, derler2011modeling, JensenChangLee11_ModelBasedDesignMethodologyForCyberPhysicalSystems, broy2012specification}. However, these approaches cover software and electric/ electronic parts of CPS, but not mechanical parts and behaviour. \cite{drave2020modeling} introduces one of the first approaches to enable a holistic view of a CPS.

For \textbf{Diagnosis of CPS} two different approaches stand out. First of all, \textit{expert-based diagnosis systems}, as first defined in~\cite{Mycin1984}, use rules to encode the system's (faulty) behaviour. While they are easy to build and to understand, they rely on expert domain knowledge and are time-consuming to build and difficult to update and maintain. \textit{Model-based diagnosis} has been used for CPS for a long time. In contrast to expert-based systems, the diagnosis model is based on a formal specification of the system's behaviour. This model is combined with observations of the system's behaviour at runtime to discover deviations from the specified behaviour. Both model and observations can be encoded with boolean/ first-principle logic \cite{reiter1987theory, de1987diagnosing}, as a discrete event system \cite{sampath1995diagnosability} or with differential equations~\cite{isermann1984process, isermann2005model}. A drawback of model-based systems is that the determination and isolation of one or multiple root causes can lead to an exponential state space. Reference~\cite{bauer2005simplifying, bauer2006model} showed how the diagnosis' inference can be transformed into a satisfiability problem and then efficiently be solved via a Satisfiability (SAT-)solver (e.g. \cite{moskewicz2001chaff}), \cite{kohl2010role} applied this to the automotive domain - a prime domain for CPS. Even though model-based systems are widely used, their models are difficult to build and understand even for domain experts, and have a rather unmanageable complexity.

Using \textbf{Machine Learning classification algorithms} for diagnosis has become very popular, especially in the medical domain. Whereas in expert-based diagnosis approaches such as~\cite{Mycin1984} expert-defined rules were used, classification algorithms learn from provided data what distinguishes faulty from non-faulty samples. These approaches can be distinguished based upon the type of the provided data. For image data, using Convolutional Neural networks~\cite{lecun1989backpropagation} is very popular. A very well-known examples for this diagnosing skin cancer~\cite{esteva2017dermatologist}, an overview of other use cases can be found under \cite{esteva2019guide}. For diagnosing CPS, these approaches have only limited applications, since alphanumeric features such as internal measurements of a CPS have to be analysed. For this kind of data, classification algorithms such as Support vector machines~\cite{cortes1995support} have been used widely since the 2000es. A well-known use case is the diagnosis of breast cancer (e.g. \cite{liu2003diagnosing}, \cite{akay2009support}). These algorithms have also been used for diagnosing machines, e.g. \cite{li2007improve}. While - given sufficient training data - these algorithms diagnose with near-human or better accuracy, they lack explainability of their results.

\textbf{Representation Learning} approaches are on the rise for explaining system behaviour. Restricted Boltzmann Machines (RBM, ~\cite{HintonSejnowski86, le2008}) are widely used for learning feature representations with newest works focusing on the explainability. \cite{abdollahi2016} proposed a RBM consisting of additional explainability units in the visible layer for recommender systems. They defined a joint distribution over visible and hidden units and a conditional distribution on explainability scores in what they called a conditional RBM. In the diagnosis field, \cite{shao2016} extracted features for motor fault diagnosis using stacked RBM~\cite{smolensky86}, i.e. Deep Belief Networks. \cite{liao2016} enhances RBM for failure diagnosis with an additional regularisation term to maintain features relevant for the health of a system. In these works, however, RBM are only used for identifying the most relevant features; explainability and structure within the feature space are not addressed.

\textbf{Bayesian Network}. In this paper, we focus on Bayesian Networks due to their better interpretability compared to other machine learning algorithms. A Bayesian network is a directed acyclic graph representing causal relationship between variables using the Bayes rule~\cite{bayes1763}. Bayesian networks were first defined by~\cite{pearl1982reverend, kim1983computational} and used for inference in tree networks. \cite{pearl1988probabilistic} extended the approach by including continuous random variables,~\cite{heckerman1995learning} combined statistical data and encoded knowledge. Bayesian Networks have been used in industry domains since the 1980s for diagnosing systems \cite{kim1983computational, heckerman1990probabilistic, breese1996decision}. The advantages of Bayesian networks are that built models and the connections of its elements can be visualised as easy understandable graphs. Additionally, the approach offers to incorporate uncertainty and compare different solutions. However, Bayesian Networks face some challenges. They have high computational costs when updating variables, as a variable's whole branch including all other dependent variables have to be updated as well. This update can extend up to the whole net, if the network has no separation. 

\textbf{Structure Learning} is the key to obtain relevant Bayesian Networks. For small scale problems domain experts can design the graph manually using their knowledge of the system, yet this approach is unpractical when learning larger networks. \textit{Constraint-based}, \textit{score-based} or \textit{hybrid} algorithms have been developed to tackle the challenge of learning Bayesian Networks. The former is typically illustrated by the PC-Algorithm \cite{pcAlg} and its modern implementation, the stable PC-Algorithm \cite{pcStableAlg}. It constructs the Bayesian graph by finding independent variables given a separating set, thus learning which variables are dependent on which, before setting arc directions. \textit{Score-based} algorithms are represented with Greedy Search \cite{GS1} and Fast Greedy Search\cite{GS2}, implementing specific modifications in best scoring graphs from previous iteration. Genetic Algorithms are also used for structural learning using a \textit{score-based} approach \cite{GASL}. Finally, \textit{hybrid} algorithms try to take advantage of both approaches, using a \textit{restrict and maximize} philosophy. Most well-known contenders of this categories include Max-Min Hill Climbing (MMHC) \cite{MMHC} and H$^2$PC \cite{H2PC}. 
All of these algorithms are usually computationally demanding and \cite{ScanagattaPaper} proposed a \textit{constraint-based} algorithm to apply the learning problem of Bayesian Networks to thousands of variables without prior expert knowledge.

\textbf{Genetic algorithms} (GA) are a common approach for algorithmic optimization. They are inspired by natural evolution. Each individual of a given population is ranked by a defined fitness function selecting only the best ones for further reproduction in a subsequent generation until convergence towards a defined optimum. Genetic Algorithms were principally defined in~\cite{fraser1970computer} and popularised by \cite{holland1975adaptation, holland1992adaptation}. They have been used for a long time to optimise machine learning algorithms. This is done by using the  GA's fitness function to maximise the relevant aspect or parameter of the machine learning algorithm. Examples for this are self-modifying weights of a recurrent neural network \cite{schmidhuber1993self}, learning the loss function of neural networks~\cite{bengio1994use}, modifying the loss function of a clustering algorithm~\cite{demiriz1999semi}, tuning a neural network's structure and parameter~\cite{leung2003tuning} or the structural learning of a Bayesian Network~\cite{Larranaga96,contaldi2019bayesian}. 

\section{Contribution} \label{sec:contribution}

\subsection{Overview}

In this section we detail our two-fold algorithm: the first part, Sec.~\ref{subsec:strucLearning}, builds a Bayesian Network for both observed features and failures. In the second part, Sec.~\ref{subsec:minimising}, we use a genetic algorithm to minimise the previously built Bayesian network by selecting the best nodes and vertices for explaining a specific failure. 

Contrary to other approaches, we learn a more general Bayesian structure using all available information before particularizing it for a specific problem instead of performing feature selection before the structural learning phase. The learning of the graph in the first step of our algorithm as well as minimising the graph in the second step are performed for a specific, given failure. The computationally heavy part in the structural learning phase and scores for the genetic algorithm can be parallelized. Hence, our algorithm is highly \textit{scalable}.

The results of the algorithm offer a more \textit{efficient} diagnosis for CPS in two ways. Firstly, the genetic algorithm generates a reduced fault pattern for a failure. Having reduced fault patterns reduce the number of states that have to be checked when inferring for the root cause of a failure and thus helps avoiding a possible state explosion problem. It also allows deploying the reduced model on CPS with limited computing power. Additionally, our approach offers saving computing resources as well as time since the necessary continuous update of the Bayesian network as of daily data updates only has to be done on the reduced network.

\subsection{Data Preprocessing and preparation}

First, we convert all non-categorical features of the dataset into discrete ordinal features as we only consider discrete probability distributions for the Bayesian Network. Hence, we use a simple binning based on quantiles with continuous values mapped to the index of their corresponding quantile with the quantiles amount a definable hyper-parameter. Non-numerical values are encoded with Boolean formulas and ordinal numbers.

\subsection{Structural learning of a failure's root cause} \label{subsec:strucLearning}

We build a Bayesian Network to learn causal relationships between variables before selection. Instead of having each variable depending on all others, we reduce the set of the variables that can influence a given variable to a parent subset. Learning the optimal Bayesian Network for a given dataset is NP-hard due to state explosion. Thus, we need to estimate the structure using a method scalable to consequent datasets. 

We approximate the Bayesian Information Criteria (BIC, \cite{schwarz1978estimating}) by using $BIC^*$ \cite{ScanagattaPaper} to measure the relevance of a set of candidate features $S$ being the parent set of a given variable $X_i$. This criterion serves as metrics to evaluate the likelihood of a set of vertices. We can compute a list of candidate parent sets for each feature $X_i$ from our data set. Given that we already have $BIC$ values for $S_1$ and $S_2$, we can use already computed scores to compute the $BIC^*$ value for a candidate $S = S_1 \cup S_2$. Thus, $BIC^*$ is used for evaluating the large number of possible candidates.

To generate the lists of candidates for each node, we use the independence selection algorithm of \cite{ScanagattaPaper}. The parent set is stored in two lists: one consisting of all explored parents and a second list for the unexplored parent set. The algorithm returns a list of the explored potential parents for a node $X_i$ with their score. 

The algorithm can be parallelized allowing an evaluation of several different candidates simultaneously. Once given the list of parent set candidates for all the variables of the dataset, i.e. all the network's nodes, we can build the network.

In the next step, parents for each node are picked given the constraints of a directed acyclic graph (DAG). Picking parents for a node sequentially introduces an anchoring bias: while the first nodes can select from all parents, later nodes are limited in their selection. To avoid this, we propose a new selection policy taking into account the order of the nodes in relation to the investigated label nodes, the failures. This policy is based on the idea that the root cause of faults, i.e. parents close to the label, are the main goal of creating our graph. So intuitively, we create the graph around the label nodes focusing on the best explanation of a given label node (in contrast to our approach~\cite{ScanagattaPaper} maximizes the graph according to BIC).

Nodes that are connected directly to the label pick first by order of relation. For instance, a second order linked variable will be picked only after all the first order variables have picked their parent sets. Variables that are separated from the label pick their parent set in a random order afterwards.

\begin{algorithm}
	\caption{Parent Set Selection} \label{alg:par-set-selection}
	\begin{algorithmic}[h!]
		\REQUIRE Parent set candidates for each node and specific $label$
		\ENSURE Parent sets for each node

		\STATE Initialize $parents$ as empty
		\STATE Initialize $open$ with $label$

		\WHILE{$open$ is not empty}
			\STATE Pop the first element of $open$ as $X$

			\WHILE{Candidate list for $X$ is not empty}
				\STATE Pop the best candidate $C$ for $X$ from candidate list

				\IF{$X$ is not a descendant of $C$}
					\STATE Accept $C$ as $X$ parent set, store it in $parents$
					\STATE Add $C$ nodes not having a parent set yet to $open$
					\STATE Break
				\ENDIF
			\ENDWHILE

			\IF{no candidate was selected}
				\STATE set $X$ parents as $\emptyset$
			\ENDIF
		\ENDWHILE

		\STATE Pick parents for the remaining nodes in random order

		\RETURN $parents$

	\end{algorithmic}
\end{algorithm}

Alg.~\ref{alg:par-set-selection} details the learning of a DAG with each node \textit{X} having an associated parent set so that the graph remains acyclic. We then can use this general architecture to extract a smaller topology that determines potential root causes of our label nodes.

\subsection{Minimising a failure's fault pattern} \label{subsec:minimising}

Given the previously learned graph, our goal is now to select a subgraph $G$ with feature nodes $E$ maximizing our ability to explain a specific label (or defect) $D$. Ideally, given all the values of the variables in the ensemble we want to be able to label $D$ correctly.

We denote the usual entropy of a discrete random variable $X$ as $H(X)$, its conditional variant for discrete random variables $X$ and $Y$ is defined as $H(X|Y)$. The discrete probabilities derived here can be estimated empirically using the maximum likelihood estimator. We also define the mutual information between two discrete random variable $X$ and $Y$ as $I(X,Y)= H(X) - H(X|Y)$ representing the amount of information shared between $X$ and $Y$ using properties from $H(X|Y)$. This conditional entropy is equal to $H(X)$ if $X$ and $Y$ are independent. It is equal to 0 if the knowledge of $Y$ gives us the whole knowledge of $X$, meaning that $X$ can be derived in a deterministic manner using the value of $Y$. This metric is often used as a distance in clustering problems and thus normalized in a symmetrical fashion \cite{mutualInf}.

In our case, since we are interested in asymetrical parent/ child relationships, we apply the \textit{Uncertainty Coefficient} defined as $U(X,Y) = \frac{I(X,Y)}{H(X)}$ \cite{UncertaintyCoeff}. This metric represents how much of $X$ can be predicted given $Y$ with $0$ in case of independent $X$ and $Y$ and $1$ for a deterministic relationship. We can now generalize this definition for $U(D,E)$ to compare how good the different sets $E$ can explain our label node $D$.

In a Bayesian Network for a given node anything outside its Markov blanket is independent and hence not relevant for the label~\cite{pearl1988probabilistic}. However, since our graph is generated based on imperfect data with neither expert nor previous knowledge, we assume that some residual information can still be grasped within higher order dependencies.
 
On top of the global information we gain from $U(D,E)$, it is also relevant to consider local information which we denote for the remainder of this section as $L(G)$. We need to consider $L(G)$ to ensure that the local inference capabilities of our model are still adequate in case some symptoms cannot be observed. Looking only at the label can lead the fitness function to ignore some key connections between features. This would be detrimental in case of key features being unobserved, as the reduced model then would not be able to predict it accurately. That means we should also consider local predicting capabilities for each feature of the reduced model, thus making the model more robust and adaptable. 

Thus, we add a term $L(G)$ to the fitness function considering the choice of kept vertices to ensure that each node still retains its relevant parent. This term balances with the global $U(D,E)$ to add some local insights to the fitness function. With $\Pi(Y)$ being the parent set of $Y$, we define this indicator as follows:
\begin{equation*}
L(G) = \frac{1}{|E|} \sum_{Y \in E}{U(Y, \Pi(Y) \cap E)}
\end{equation*}

From the definition of $U(D,E)$ we can see that this value increases for larger sets $E$ with more states. Hence, the defined genetic algorithm would tend to increase the number of selected features regardless of their importance and thus potentially leading to an overfitting. To address this, we use a $L2$-regularization term on the number of states from $E$. This term introduces two hyper-parameters depending on the dataset, the characteristic state number $\tau$ denoting the expected number of states and the intensity of the regularization $C$ (typically $10^{-3}$) for balancing the regularisation term. We define the regularization term as:
\begin{equation*}
R(E) = C\max(0,(\frac{|E| - \tau}{\tau}))^2
\end{equation*}

Although, $L2$ regularization leads to smaller weights or in our case fewer states, it allows us to penalize states directly connected to the target far less than the ones of higher order. With the input of additional generations in the genetic algorithm, the number of total states increases. Therefore, we use the $L2$ regularisation to ensure that only states adding relevant information are kept in the reduced model, since later added states will be penalized more as they will be further away from the target. 

Finally, using the previously introduced terms, we can define our fitness function $F(G)$ as:
\begin{equation*}
F(G) = U(D,E) + L(G) - R(E)
\end{equation*}

Thus, we search the subgraph $G$ maximising $F(G)$ by exploring available graph topologies, starting from the label node and going up its parents. The genetic algorithm then explores the possible topologies and retains the best ones. This method is scalable since the evaluation of the fitness function is parallelized. 

\begin{algorithm}
	\caption{Root Cause Extraction} \label{alg:root-cause-extraction}
	\begin{algorithmic}[h!]
		\REQUIRE DAG $G$, $label$, $max\_gen$, $K$, $patience$, $plateau$
		\ENSURE A reduced DAG from $label$'s parents

		\STATE Initialize $generation$ as empty
		\STATE Initialize $stagnation$ and $gen\_number$ with 1

		\FOR{all combination $E$ of $label$'s parents in $G$}
			\STATE Add $F(label,E)$ to $generation$
		\ENDFOR

		\STATE Select the $K$ fittest individuals as $best$

		\WHILE{$stagnation < patience$ and $gen\_number < max\_gen$}
			\STATE Set $generation$ as empty
			\STATE Add $best$ individ. from previous to $generation$
			\STATE Breed $best$ individuals and add them to $generation$
			\STATE Select new K fittest individuals from $generation$
			\IF{difference between top individuals $< plateau$}
				\STATE Increment $stagnation$
			\ENDIF
			\STATE Increment $gen\_number$
		\ENDWHILE
		\RETURN $generation$ fittest individual

	\end{algorithmic}
\end{algorithm}

This algorithm allows us to test a wide range of different extracted graphs depending on hyper-parameter tuning. It can be adapted using different breeding methods such as exhaustive search (given a narrow network or high computational power) or random mutation.

Hence, we aim for first learning relationships of the whole CPS and then select the relevant ones for a given incident. The fitness function $F$ in Alg.~\ref{alg:root-cause-extraction} measures separatibility between classes given the parents. When a feature allows to separate a significant number of positive samples from the other overwhelming majority of negative ones, the $U$ term for the fitness will increase greatly. This approach does not rely on a balance of samples thus making it suitable for identifying potential root causes. It is also worth mentioning that $U$ is the non-symmetrized version of symmetrical uncertainty which has been shown as relevant for high-dimensional feature selection \cite{symU}.
Our approach first learns a global representation of the data, meaning that all vertices will be considered and weighted regardless of their affiliation with the label. It is ensured that the edges of the reduced model are relevant according to the BIC-criteria as they were picked by a global model in the first place against many others. Thus, we ensure the relevance of our final vertices.

\subsection{Sequential failures} \label{subsec:sequentialFailure}

Thus far, we focused on single failures. For CPS and humans, however, multiple failures or illnesses happen quite often. These events can either happen simultaneously at the same time or sequentially. An example for a sequential failure would be a defect connection between CPS' components leading to a time-out of the receiving component leading to a failure since the component could not execute a required task within a deadline. In this section, we show how our methodology can cover sequential failures, multiple failures will be covered in this paper's outlook.

Since during Alg.~\ref{alg:par-set-selection} symptoms and failures are both treated as features, we can learn an existing dependency between 2 distinct failures $F_i$ and $F_j$, with $F_i$ occurring before $F_j$. In terms of our algorithm, $F_i$ would then be the parent node for $F_j$. This situation can be difficult to model using traditional classifiers since the state of $F_i$ can be unknown, thus being unavailable for inference since it then was not part of neither training nor test set. On the contrary, a Bayesian Model can use the previously learnt conditional distribution to provide insights on unobservable nodes. 

In contrast to these classifiers, during the inference phase of Bayesian Networks, hidden variables are regularly used. Regarding failure $F_j$, this means that the Bayesian network infers $F_i$'s state using its parent nodes dependencies when predicting our target $F_j$. 

Hence, this helps identifying sequential faults and offers an easy interpretable visualisation of their dependencies and interactions.

\section{Evaluation} \label{sec:evaluation}

We evaluated our approach on two use cases. The first one is a diagnosis of an automotive component, the second a medical diagnosis of humans. We chose medical diagnosis since our main challenges imbalanced data and interpretability hold for the medical domain. For the medical use case, we will use the terms illness instead of failure and symptoms for features.

The data for both use cases is heavily imbalanced and consists of hundreds or thousands of features of which only a small amount indicates a complex fault pattern. The data for the automotive use case is based on development data which we cannot publish due to confidentiality. Medical data is rarely publicly available, so we chose to generate a data set for the evaluation based on a public dataset that has relations between symptoms and diseases. 

We mentioned in Sec.~\ref{subsec:challenges} that for imbalanced data sets many feature-based selection techniques regard the few faulty ones as noise. Additionally, the results of these algorithms lack interpretability. Hence, the main criterion for our evaluation is having a comparable accuracy and an improved interpretability. 

The evaluation shows that our approach overcomes the mentioned problems while returning a comparable accuracy.

\subsection{Automotive Use Case}

Our first use case is the root cause analysis of a failure of a mechanical component of the powertrain occurring during its development. In Sec.~\ref{subsec:challenges} we detailed why measurement data for mechanical parts is scarce. Hence, we need to link all available data for an analysis. This leads to a sparse data set with complex fault patterns which initially motivated our approach of a more explainable representation. 

We built two classes, one with vehicles having the failure (the label) and the other with similar vehicles not having this failure. As mentioned, the data is heavily imbalanced, since specific failures only came up in very few vehicles.

\begin{figure}[ht!]
    \centering
    \begin{subfigure}{.48\textwidth}
        \centering
        \includegraphics[width=0.75\textwidth]{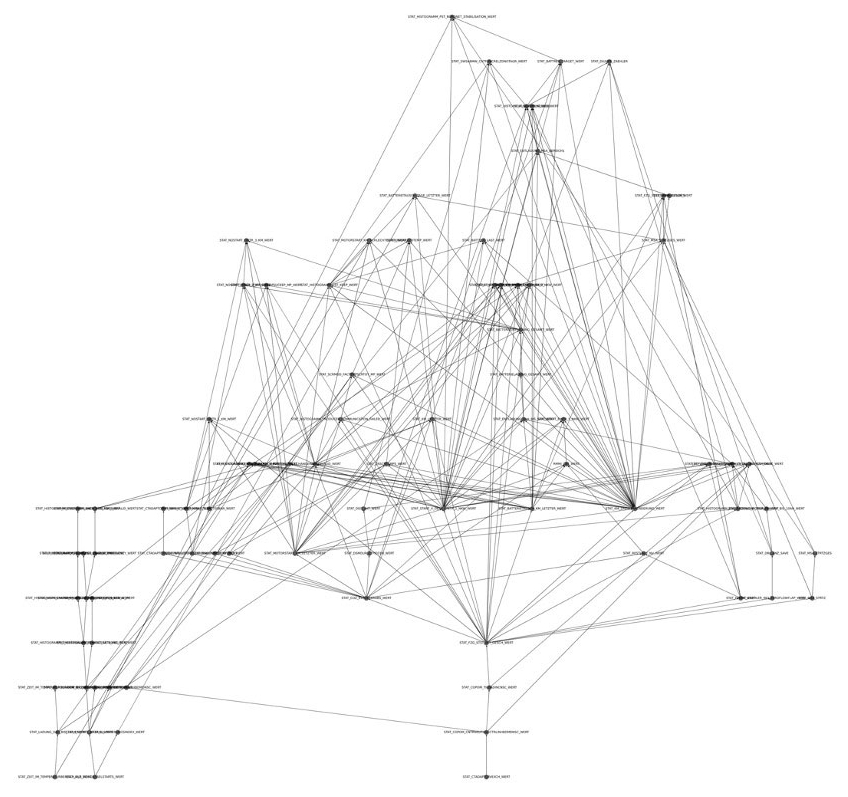}
        \caption{Extract of learnt Bayesian Network (68 nodes, 99 vertices)}
    \end{subfigure}
    \begin{subfigure}{.49\textwidth}
        \centering
        \includegraphics[width=0.75\textwidth]{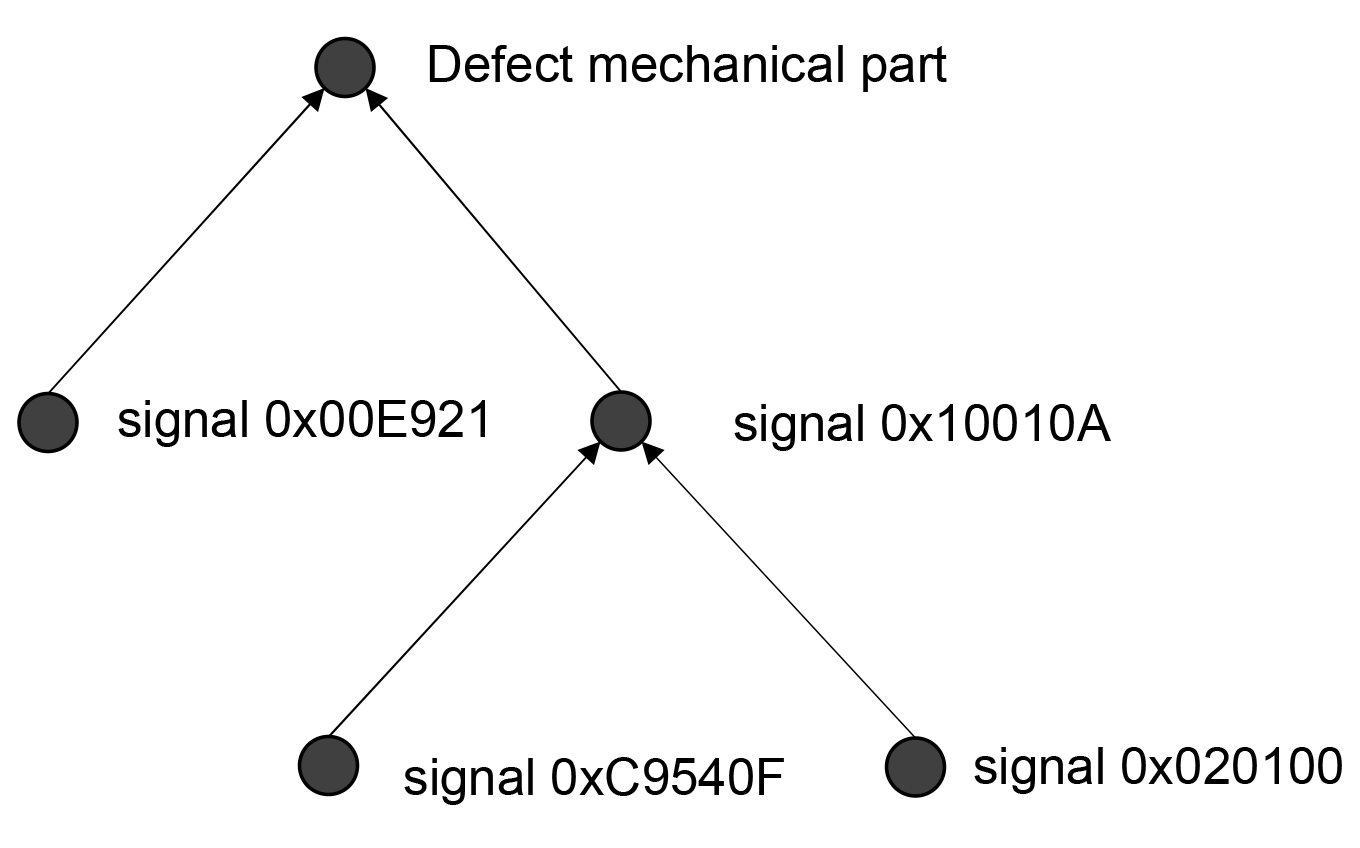}
        \caption{Reduced Bayesian Network (5 nodes, 4 vertices)}
    \end{subfigure}
    \caption{Learnt representation for the automotive use case}\label{fig:autorep}
\end{figure}

From the initial $3000$+ available features, we built a Bayesian Network numbering around $1800$ nodes. Using the genetic algorithm, we reduced it down to four features for a specific failure, which we then verified with domain experts. To ease understanding, Fig.~\ref{fig:autorep} shows only an excerpt of the learnt Bayesian Network and short type information for the features in the reduced network. 

\subsection{Medical Diagnosis Use Case}

As basis for our second use case, we chose a medical data set on Kaggle~\cite{medDataset}. The original dataset contains $41$ different pathologies as target classes linked to $132$ different symptoms. We generated $500$ patients for each pathology. Our dataset uses a probabilistic approach with each symptom having an appearance probability depending on the type of pathology simulated.

\begin{figure} [ht!]
    \centering
    \begin{subfigure}{.48\textwidth}
        \centering
        \includegraphics[width=0.75\textwidth]{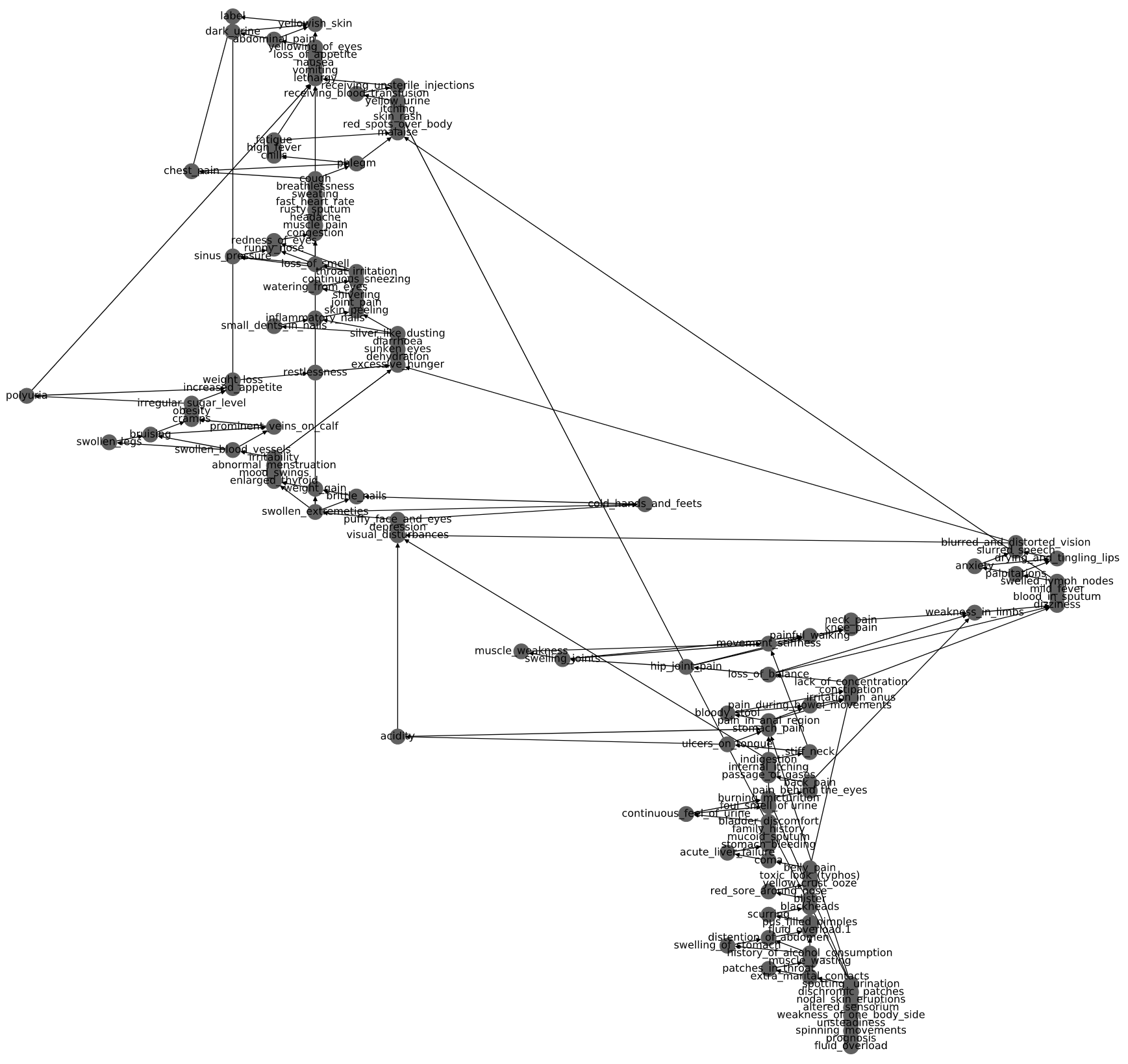}
        \caption{Fully learnt Bayesian Network ($133$ nodes, $192$ vertices)}
    \end{subfigure}
    \begin{subfigure}{.49\textwidth}
        \centering
        \includegraphics[width=0.75\textwidth]{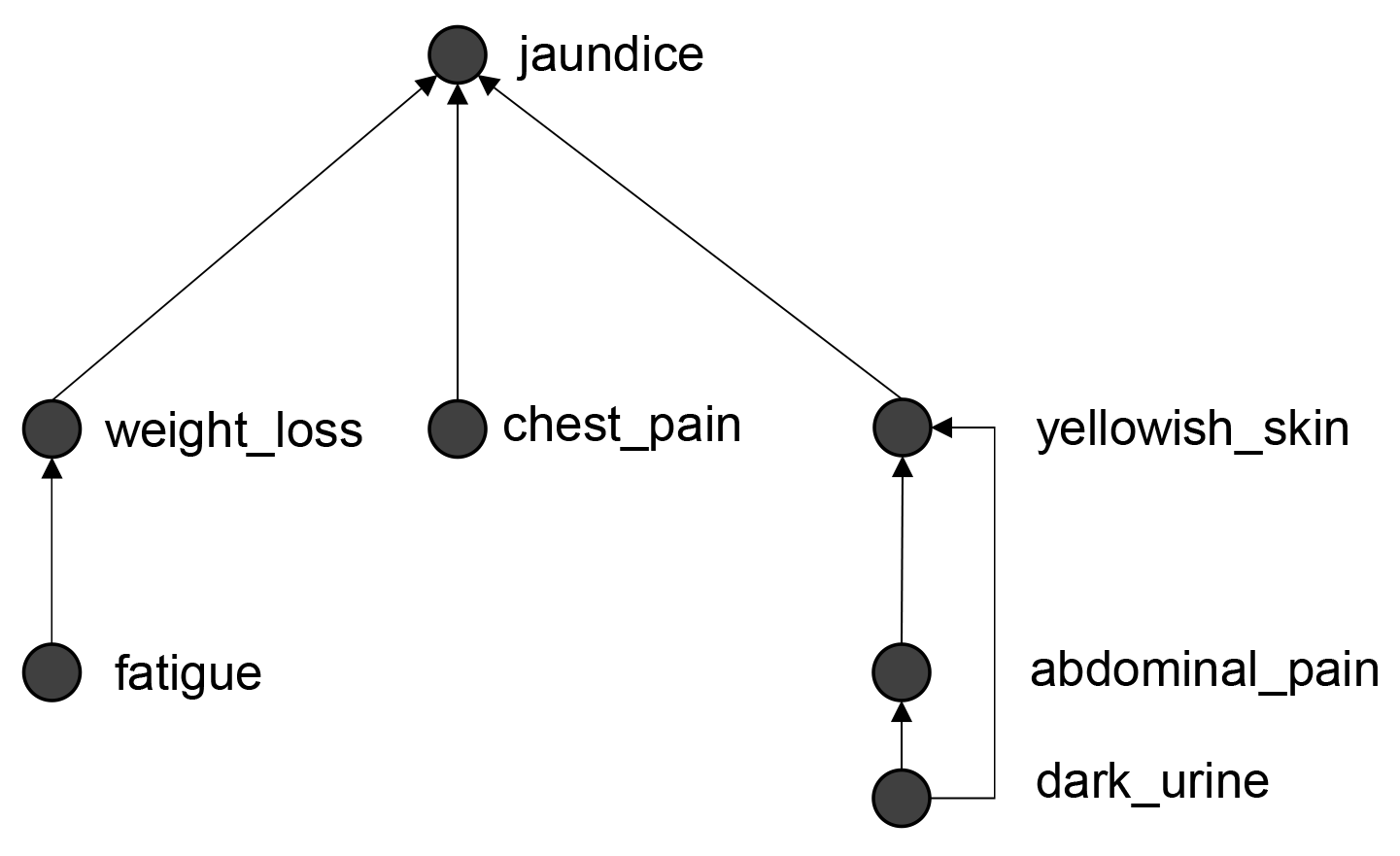}
        \caption{Reduced Bayesian Network ($7$ nodes, $7$ vertices)}
    \end{subfigure}
    \caption{Learnt representations for the medical data}\label{fig:medrep}
\end{figure}

From this data set we chose the \textit{jaundice} disease as label and generated a dataset of 20'500 patients with diverse symptoms. This dataset is heavily imbalanced (less than 2\% positive samples) and contains $132$ different symptoms or features. Fig.~\ref{fig:medrep} shows the learnt models. This representation contains seven out of $132$ features. More importantly, all six features are among \textit{jaundice} symptoms. 

We now compare our approach to the common classifiers such as Decision Tree~\cite{breiman1984classification}, Random Forest~\cite{ho1995random} and Gradient Tree Boosting~\cite{chen2016xgboost}. Additionally, we fed the nodes of the reduced Bayesian Network in a constrained decision tree with a maximum depth of five. By using the constrained decision tree with its depth limit we can check on the accuracy whether we chose the right features.

Table~\ref{table:results} shows the classifiers' results for the positive class, its precision, recall and F1-score. Precision represents the amount of relevant positive items among the predicted positive items. Recall, or sensitivity, on the other hand, depicts how many among the relevant positive items are predicted positive by our model. F1-score mixes these two indicators for an overall performance estimation and is typically used to select the prefered classification method. For diagnosis use cases, precision or specificity is very interesting, although in our use cases the imbalanced ratio makes it hard to discriminate on the latter.

\begin{table} [ht!]
    \centering
    \begin{tabular}{l| c | c | c | c}
        Model & Precision & Sensitivity & Specificity & F1-Score \\
        \hline
        Decision Tree & 0.67 & 0.69 & 0.99 & 0.68 \\ 
        Random Forest & 0.78 & 0.68 & 0.99 & 0.73 \\ 
        XGBoost & 0.80 & 0.77 & 0.99 & 0.78\\ 
        Red. Bay. Network & 0.80 & 0.65 & 0.99 & 0.72\\
        Constr. Dec. Tree & 0.80 & 0.77 & 0.99 & 0.78 
    \end{tabular}
    \caption{Comparison of results for positive class (jaundice)} \label{table:results}
\end{table}

Regarding precision, all of the algorithms performed quite similar except decision tree which had a significantly lower value. The reduced Bayesian network seems comparable to a random forest in terms of performance, but uses a lot less features and is easier to interprete.  
The benefit of the Bayesian network and its derived constrained decision tree is the improved interpretability compared to Random Forest and XGBoost. While these models use 100 different estimators of various depth, the reduced graph of the Bayesian network in contrast consists of 6 (of which only 3 are actually used for jaundice prediction due to Markov separation) out of the initial 132 features. 

\subsection{Sequential diagnosis}

In this section, we show how our algorithm can learn the relation between sequentially occuring illnesses on two related medical diseases. \textit{Gilbert's syndrome} is a harmless liver condition due to a genetic mutation. One of its most frequent symptoms is our previous example of \textit{jaundice}, thus making this a good candidate for sequential diagnosis.  

A patient is having an increased chance of suffering from Gilbert's syndrome when she is already diagnosed with \textit{jaundice} as well as some other characteristic symptoms such as \textit{cough}, \textit{high fever} or a \textit{yellowish skin}. This means that a previous diagnosis of jaundice or having a priori knowledge of jaundice causes is a key information for the diagnosis of this new disease. 

\begin{figure}[ht!]
    \centering
    \includegraphics[width=0.45\textwidth]{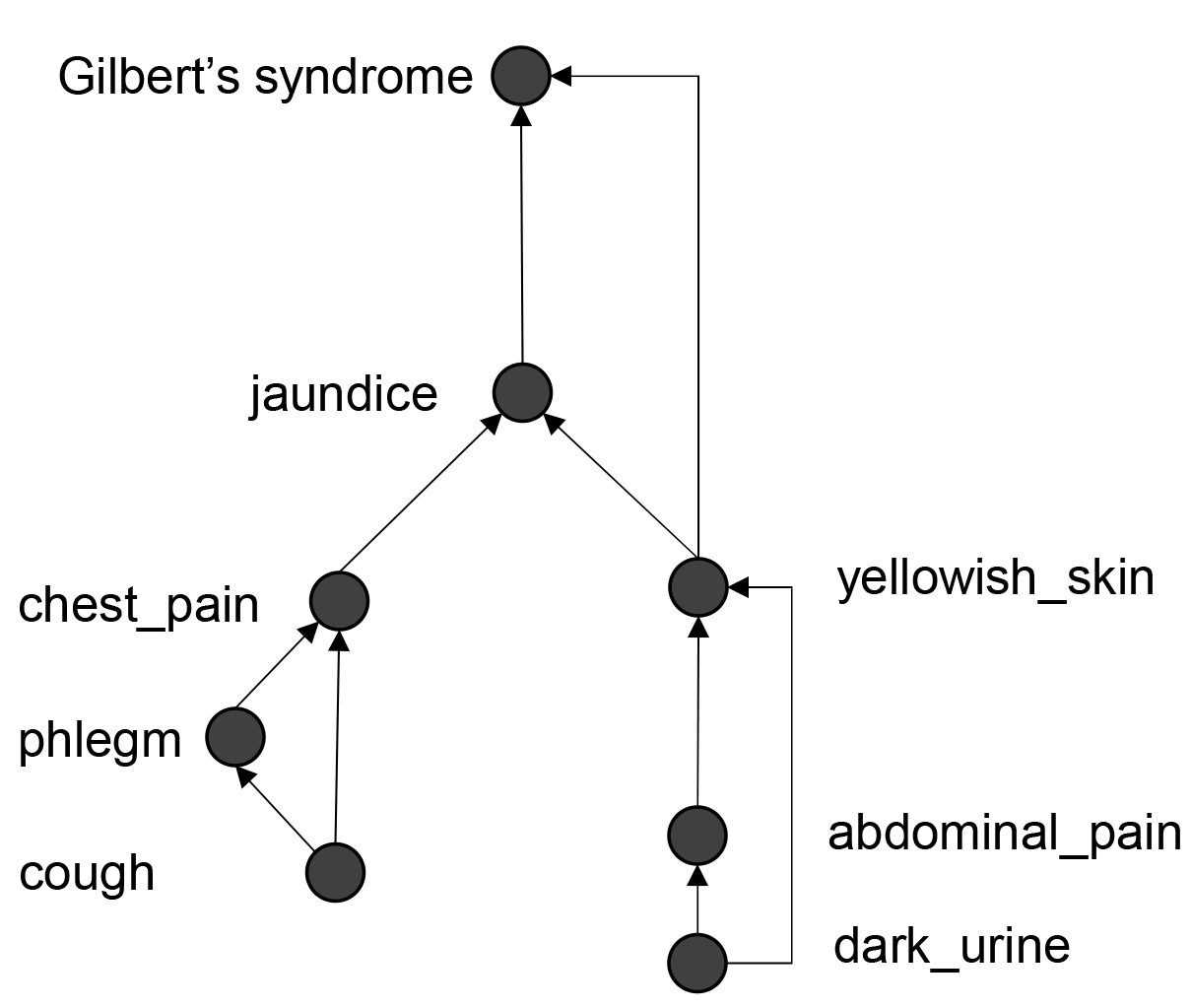}
    \caption{Reduced learnt representation for Gilbert's syndrome (8 nodes, 10 vertices)}\label{fig:newdisease}
\end{figure}

Using the incremental learning procedure, we add the new label Gilbert's syndrome to the previously learnt Bayesian network for the medical use case shown in Fig.~\ref{fig:medrep} without need for recomputing the existing candidate sets. We then perform a root cause analysis to compute the reduced graph Fig.~\ref{fig:newdisease} for the new disease. 

This graph shows that our approach managed to identify the mentioned key symptoms of Gilbert's syndrome as well as learning a relation between the two diseases. Additionally, for several jaundice symptoms a transitive link to the label Gilbert's syndrome was established. This might be helpful on diagnosing Gilbert's syndrome especially when some symptoms are not yet observable. Regarding inference, the jaundice node is considered unobservable, i. e. we do not know if the patient has jaundice and only infer based upon the symptoms.

We can now compare the performance of our reduced Bayesian model with other classifiers which do not have jaundice as input. It is also interesting to note that the Bayesian Model can also infer the most probable state of this hidden variable through the inference process. Results are displayed in table~\ref{table:newresults}.

\begin{table} [ht!]
    \centering
    \begin{tabular}{l| c | c | c | c}
        Model & Precision & Sensitivity & Specificity & F1-Score \\
        \hline
        Decision Tree & 0.24 & 0.23 & 0.98 & 0.24 \\
        Random Forest & 0.52 & 0.21 & 0.98 & 0.30\\
        XGBoost & 0.59 & 0.27 & 0.98 & 0.37\\
        Red. Bay. Network & 0.53 & 0.35 & 0.99 & 0.42 \\
        Constr. Dec. Tree & 0.53 & 0.30 & 0.98 & 0.38
    \end{tabular}
    \caption{Comparison of results for positive class (Gilbert's syndrome)} \label{table:newresults}
\end{table}

This table shows that our reduced Bayesian model has a comparable performance with other classifiers for sequential faults. 

Conclusively, we showed how our algorithm gives valuable and reliable insights for the root cause analysis of single and sequential faults even without a prior knowledge of the system. Additionally, our algorithm is capable of identifying key features or symptoms given failure(s) or illness(es) in large datasets.


\section{Conclusion and Future Work} \label{sec:conclusion} 

We presented a scalable algorithm capable of learning a structured, optimised diagnosis model for cyber-physical systems with comparable accuracy to standard approaches but higher interpretability.  

With its optimised and separated structure, the learnt diagnosis model can be easily understood and provides concise, consistent and accurate diagnosis results.

Additionally, the learned model can be easily integrated into the workflow of diagnosis engineers providing benefits such as an improved model maintenance or by easily integrating further knowledge to the model in form of boolean formulas. Moreover, they can easily add new failures and faults to the existing model by computing the fault's parents using the described methodologies and then search for the failure's root cause(s). This allows for continuous, iterative development, incremental model upgrades and thus an improved maintainability.

Given these points, we are convinced that our approach helps domain experts and outweighs the inherent mentioned disadvantages of Bayesian networks. Additionally, we can use the learned, reduced model for other methodologies such as supervised learning algorithms or "traditional" model-based diagnosis approaches.

Regarding future work, we are planning to extend our approach for diagnosing multiple failures. In contrast to the shown sequential faults, these failures are occurring at the same time, i. e. into a multi-label classification problem.

Finally, since we can annotate the model with repair measures, their costs and repair time as well as chance of success, we can create a real-options model~\cite{black1973pricing} for optimising repair measures~\cite{Haddad2011}: for example, changing a specific part of a CPS can be cheaper than changing the whole CPS but requires more time and has a different probability of success.

\bibliographystyle{IEEEtran}
\bibliography{sources}

\end{document}